\documentclass{article}
\usepackage{ijcai22}

\usepackage{times}

\usepackage{soul}
\usepackage{url}
\usepackage[hidelinks]{hyperref}
\usepackage[utf8]{inputenc}
\usepackage[small]{caption}
\usepackage{graphicx}
\usepackage{amsmath}
\usepackage{booktabs}
\usepackage{multirow}
\usepackage{xcolor}
\usepackage{dsfont}

\title{From Seeing to Moving: A Survey on Learning for Visual Indoor Navigation (VIN)}

\author{
Xin Ye\And
Yezhou Yang\\
\affiliations
SCAI, Arizona State University, USA\\
\emails
\{xinye1, yz.yang\}@asu.edu
}

\begin{document}

\maketitle

\begin{abstract}
Visual Indoor Navigation (VIN) task has drawn increasing attention from the data-driven machine learning communities especially with the recently reported success from learning-based methods. 
Due to the innate complexity of this task, researchers have tried approaching the problem from a variety of different angles, the full scope of which has not yet been captured within an overarching report.
This survey first summarizes the representative work of learning-based approaches for the VIN task and then identifies and discusses lingering issues impeding the VIN performance, as well as motivates future research in these key areas worth exploring for the community.
\end{abstract}

\section{Introduction}

John McCarthy, who coined the term Artificial Intelligence back in 1955 \cite{mccarthy2006proposal}, defines it as the {\it``science and engineering of making intelligent machines'', in which an
intelligent agent is a system that perceives its environment and takes actions that maximize its chances of success to achieve certain goals.} 
Visual Indoor Navigation (dubbed as VIN) 
fits exactly this very definition of an AI task,
where an intelligent agent (a.k.a. robot) is instructed to navigate towards a user-specified goal in an indoor environment based on its first-person visual observations (typically the RGB images captured by its on-board camera). It is a fundamental yet an integral task towards achieving the goal of Artificial Intelligence, which requires the agent to be able to {\it understand} its visual inputs, {\it infer} its current location, {\it reason} about the goal location, {\it plan} a trajectory, and {\it execute} an action to perform at each step. The capability of performing VIN well further enables a variety of higher level AI tasks, such as Embodied Question Answering \cite{das2018embodied} where the agent needs to navigate to a question-specified target location to gather visual information for answering questions, and Vision-and-Language Navigation \cite{anderson2018vision} in which the agent has to follow the human language instructions to navigate the indoor environments. As a result, VIN has drawn an increasing amount of research attention, and inspired a large amount of work attempting to tackle it.

Classical map-based methods for visual navigation have been studied for years \cite{bonin2008visual}. These methods explicitly decompose the navigation task into a set of sub-tasks, i.e. mapping, localization, planning and motion control. 
Although these methods have achieved a decent amount of success of the years, modular design has fundamental limitations preventing their widespread adoption.
One significant limitation is their susceptibility to sensors' noises accumulating and propagating down the pipeline from the mapper to the controller, making these algorithms less robust in complicated environments.
More importantly, they require extensive case-specific and  scenario-driven manual-engineering, making them difficult to integrate with other downstream AI tasks that have achieved superior performance with the data-driven learning methods, such as visual recognition, question answering, and scene captioning.

\begin{figure}[t!]
\centering
\includegraphics[width=0.95\columnwidth]{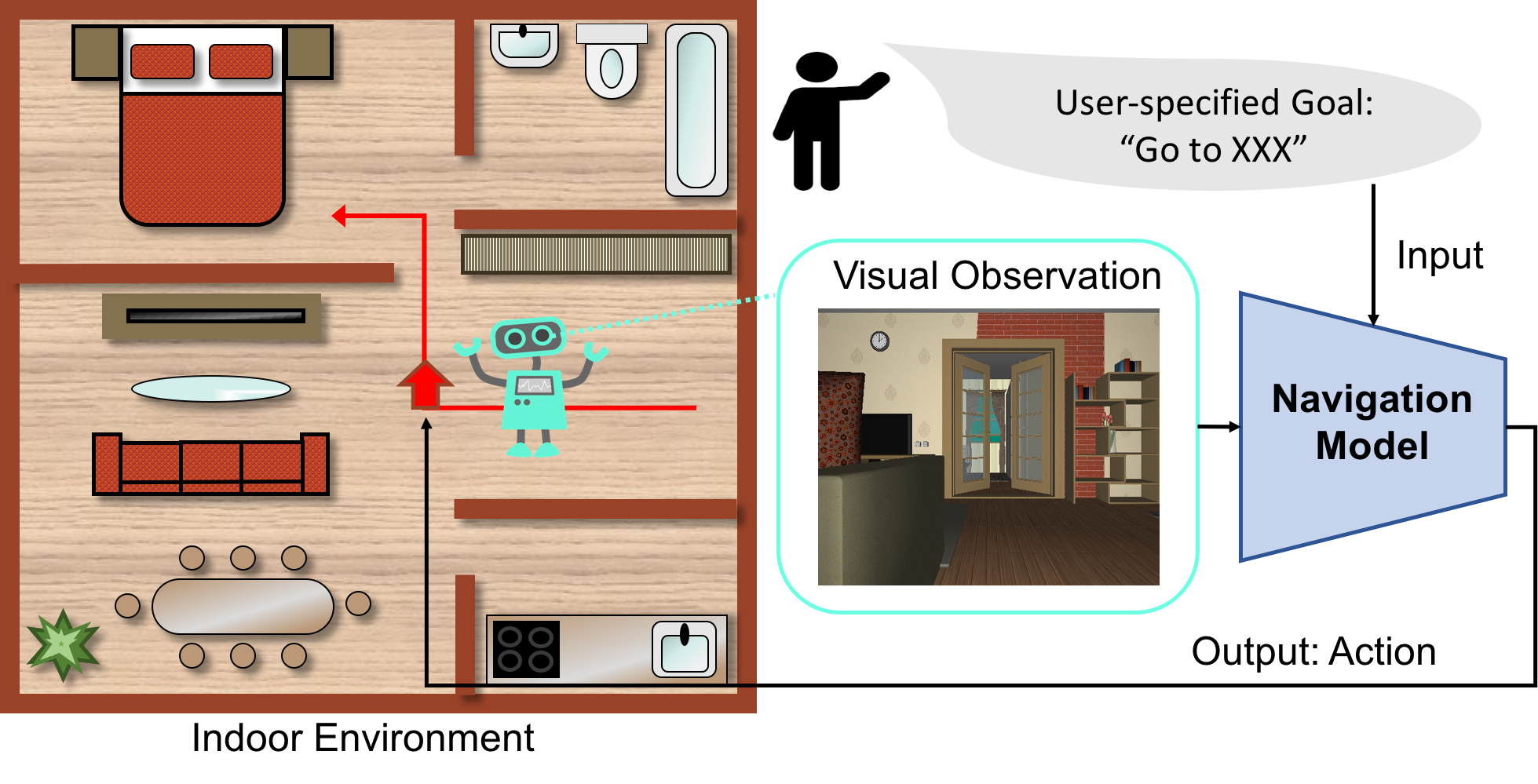}
\caption{An illustration of a learning-based method for visual indoor navigation task.}
\label{fig:VN}
\vspace{-7pt}
\end{figure}

\begin{figure*}[ht]
\centering
\begin{minipage}[c]{0.42\textwidth}
\centering
\includegraphics[width=\textwidth]{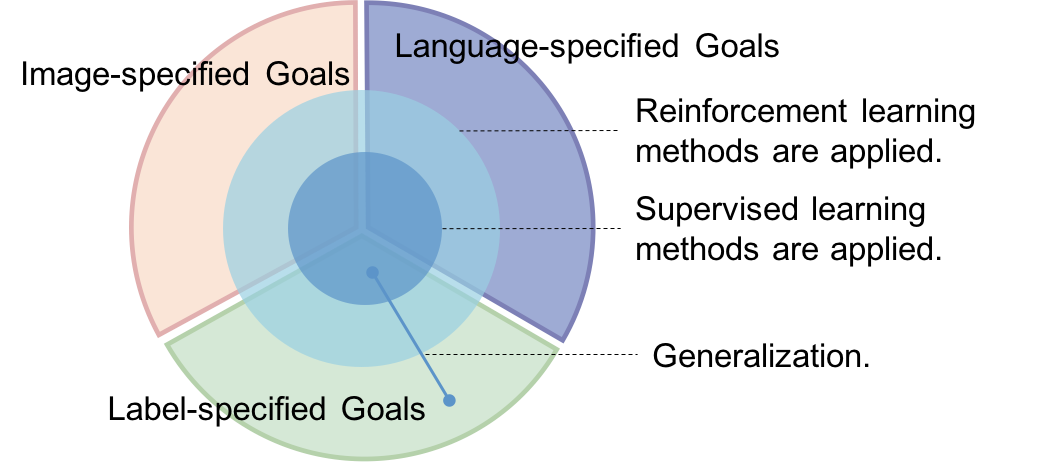}
\end{minipage}
\begin{minipage}[c]{0.53\textwidth}
\centering
\small
\begin{tabular}{|l|l|l|}
    \hline
    Tasks  & Methods & Related Work \\ \hline \hline
    \multirow{3}{*}{Label} & SL only & \cite{gupta2017cognitive}, \cite{mousavian2019visual} \\ 
    \cline{2-3}
    & RL only & \cite{druon2020visual}, \cite{yang2018visual} ...
    \\
    \cline{2-3}
    & SL + RL & \cite{chaplot2020object}, \cite{yang2019embodied}, ...
    \\
    
    \hline
    \multirow{3}{*}{Image} & SL only & \cite{savinov2018semi}, ...
    \\ 
    \cline{2-3}
    & RL only & \cite{devo2020towards}, \cite{zhu2017target}, ...
    \\
    \cline{2-3}
    & SL + RL & \cite{lv2020improving}\\
    \hline
    
    \multirow{3}{*}{Language} & SL only &  \cite{fried2018speaker}, \cite{ma2019self}, ...
    \\ 
    \cline{2-3}
    & RL only &  \cite{gordon2018iqa}, ...
     \\
    \cline{2-3}
    & SL + RL & \cite{das2018embodied}, \cite{das2018neural}, ...
    \\
    \hline
\end{tabular}
\end{minipage}
\caption{A brief summary of the VIN tasks categorized by their goals' descriptions, and the applied methods (SL: Supervised Learning based methods; RL: Reinforcement Learning based methods).}
\label{fig:summary}
\vspace{-7pt}
\end{figure*}

Observing learning-based methods' recent success in related tasks, there has been a surge of works applying them  to VIN challenges.
As shown in Figure~\ref{fig:VN}, the learning methods typically take visual inputs and user-specified goals as inputs and output an optimal action for the agent to take at each timestamp in order to achieve the user-specified goals. 
As opposed to classical methods, learning-based methods infer solutions directly from the data and as a consequence, require little manual-engineering and serve as a foundation for novel AI-driven visual navigation tasks.
 While it is promising, learning to navigate also poses challenges to tackle. For example, how to efficiently represent the visual inputs? How to reason the connection between the current observation and the user-specified goal location, especially when they are from different modalities? And, how to train the model without the ground-truth actions labeled, etc?
 Each challenge warrants extensive research efforts and overarching guiding theories are still lacking, resulting in scattered perspectives.
This survey aims to provide readers with a more holistic understanding of recent work in learning-based methods for VIN, with an attempt to provide guidance on how general theories and protocols for the field could be achieved.

Specifically, the survey categorizes the recent learning-based visual navigation work into certain high level categories.
We specify what aspects of the VIN system these works improves respectively.
We revolve the visual navigation system and discuss what is still missing to further improve the VIN performance. Lastly, we summarize the current progress of the learning methods on this task and conclude with listing the future directions of where the research can progress towards. 

\begin{figure*}[t!]
\centering
\includegraphics[width=0.9\textwidth]{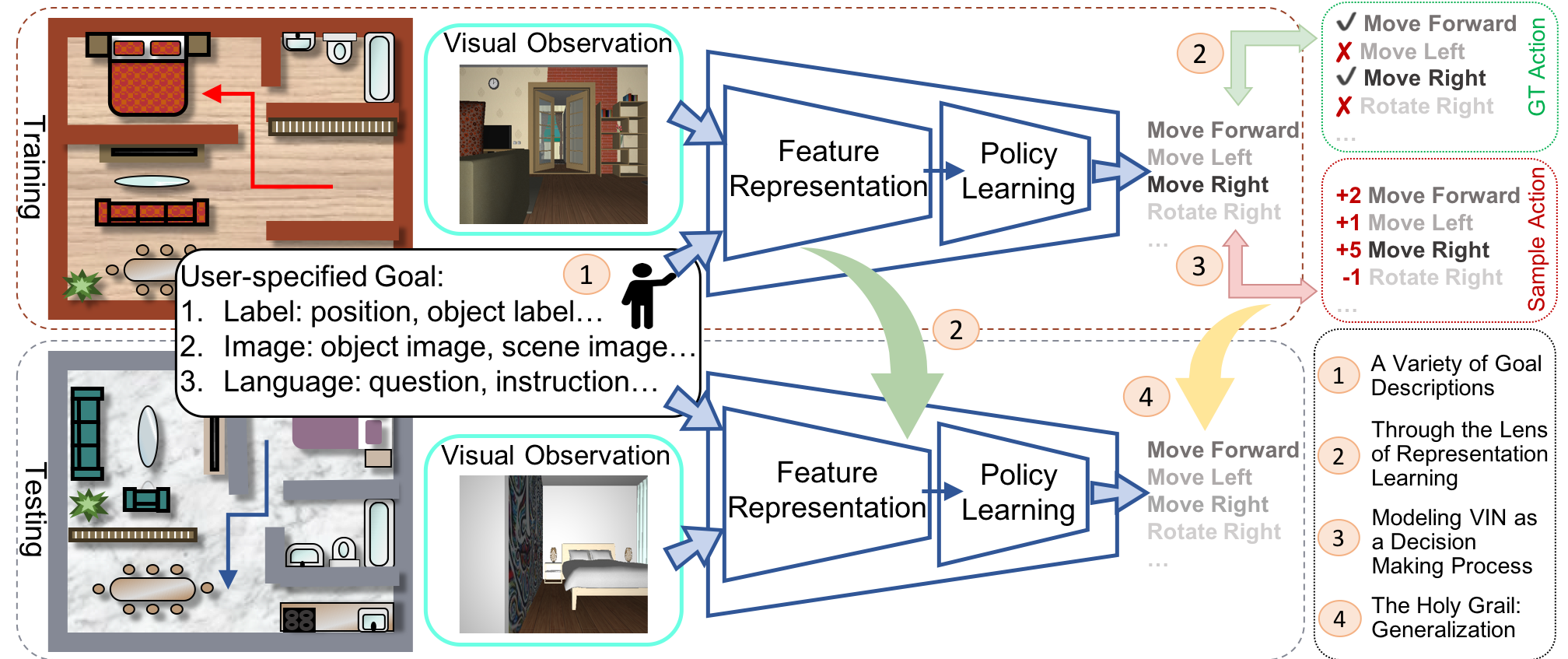}
\caption{An illustration of the recent research on VIN (Visual Indoor Navigation) task.}
\label{fig:survey}
\end{figure*}

\section{VIN Overview}
\label{sec:summary}
To start, we describe the recent approaches that address the VIN problem with the learning methods. These works are categorized as depicted in Figure~\ref{fig:summary} and reviewed in the following sections. 
\subsection{A Variety of Goal Descriptions}
\label{sec:summary_rep}
Visual observations and user-specified goals are the two inputs to learning-based visual navigation models, with tasks being defined based on the latter.
We summarize them in terms of the descriptions of the user-specified goals (as shown in Figure~\ref{fig:survey}).

{\bf Goals described by labels.} In a 2D environment or an environment where the map is known or learned, it is straightforward to specify the goal with an absolute 3D position $(x, y, z)$ that is defined in the coordinate frame of either the environment or the agent \cite{gupta2017cognitive}. However, in the first-person view navigation task, the map information is not necessary and thus specifying the absolute goal position is not efficient. 
It is more common to specify the goals with the semantic labels that can be inferred from the visual observation, such as room types or object categories in order to ask the agent to navigate to the designated rooms \cite{wu2018building,wu2019bayesian} or search target objects \cite{yang2018visual,ye2019gaple,mousavian2019visual,druon2020visual,du2020learning,chaplot2020object,wani2020multion,ye2021auxiliary}. Consider that the inconsistent interpretations of the object navigation task are emerging among the increasing number of works, \cite{batra2020objectnav} further summarizes the consensus recommendations and suggests several existing experimental platforms for this object goal navigation task. \cite{yang2019embodied} further proposed a task called Embodied Amodal Recognition (EAR) - the goal of the agent is to correctly classify, localize and segment the target occluded objects through the viewpoints collected during the agent's navigation. 

{\bf Goals described by images.} \cite{zhu2017target,savinov2018semi,watkins2019learning,devo2020towards,lv2020improving,chaplot2020neural} described the goals with the visual observations of the scenes at the goal positions.
In \cite{wu2019exploring}, the authors adopted the visual observations that contain objects as the goals to guide the agent to approach the image-indicated object, while in \cite{ye2018active}, the authors provided the target objects' images without any contextual information included.

{\bf Goals described by natural language.} Two primary tasks that take human language as navigation goals have drawn much attention. The first is the Embodied Question Answering (EQA) \cite{das2018embodied,gordon2018iqa} task. The task asks questions that require the agent to navigate in an indoor environment and collect visual information to infer an answer. More advanced variations can be found in \cite{das2018neural,yu2019multi,wijmans2019embodied}. The second is the Vision-and-Language Navigation (VLN) task proposed by \cite{anderson2018vision}, in which navigation instructions are provided in the form of natural language. Unlike other goal-driven navigation tasks, VLN task requires both the goal locations and the navigation trajectories to be aligned with the provided instructions. The task has been extensively studied in \cite{wang2018look,fried2018speaker,tan2019learning,huang2019transferable,wang2019reinforced,hong2021vln,chen2021history}.  Different from the fine-grained visuomotor instructions provided in the VLN task, \cite{qi2020reverie} introduced the REVERIE task where the agent is given a high-level natural language instruction referring to a remote object. \cite{zhu2021soon} presented the SOON task where the agent is instructed to find a thoroughly described target object inside a house. Such high-level instructions better reflect the way how humans communicate.

\subsection{Through the Lens of Representation Learning}
\label{sec:sl}
With many simulation platforms being developed for visual navigation tasks, such as AI2-THOR \cite{kolve2017ai2}, House3D \cite{wu2018building}, R2R \cite{anderson2018vision} and Habitat \cite{savva2019habitat}, the optimal trajectories for most visual navigation tasks are accessible. For example, the optimal trajectories for most goal-driven navigation tasks are the shortest paths from the agent's current locations to the goal locations where the two locations are specified by the agent's visual observations and the goal inputs, respectively. Even for the VLN task, the corresponding R2R dataset also provides desired trajectories as references. As a consequence, each visual observation $o$ together with the user-specified goal $g$ is associated with the optimal action $a$, which can serve as training data $\{(o, g, a)\}$ that enables supervised learning method to address the visual navigation problem. To be specific, 
the supervised learning method approximates the action policy $P(a|o; g)$ by  a function $f$ in which $f(o; g) = \mathds{1}(a)$ for the training data with the hope that it can generalize to the testing data. In visual navigation task, such function $f$ requires capturing strong feature representations from the visual observations as well as the user-specified goals.

Taking visual navigation as feature representation learning is common in VLN task, as the preliminary challenge of the VLN task is the cross-modal grounding of the visual observations and natural language instructions.  With the desired trajectories provided in the VLN benchmark dataset R2R, many works make effort towards learning better feature representations for VLN tasks. \cite{wang2019reinforced} presented a cross-modal matching architecture to ground language instruction on both local visual observation and global visual trajectory. \cite{hong2020language} proposed a language and visual entity relationship graph modeling relations among scene, object and directional clues. In \cite{ma2019self,zhu2020vision,huang2019transferable}, the authors proposed self-supervised auxiliary tasks to accelerate the learning of the effective feature representations.
More recently, transformer-based architectures are explored to better model the cross-modal relationships between vision and language \cite{hong2021vln} and/or long-horizon history of observations and actions \cite{chen2021history}.

In addition to the VLN task, other goal-driven visual navigation tasks also learn the feature representations in the presence of the optimal trajectories. In \cite{mousavian2019visual}, the authors evaluated various visual representations for goal-driven navigation. \cite{shen2019situational} proposed to fuse a large set of visual representations to develop diverse visual perception abilities.
Map-like environment representations are always used for navigation tasks, such as non-parametric landmark graph \cite{savinov2018semi}, top-down egocentric obstacle map \cite{gupta2017cognitive} and semantic map \cite{chaplot2020object}  built from visual observations.
The utility of the map representations on navigation performance is discussed in \cite{wani2020multion}.


\subsection{Modeling VIN as a Decision Making Process}
\label{sec:rl}

Even with the training data $\{(o, g, a)\}$, a VIN task is not simply a  supervised learning problem as the observations $\{o\}$ are neither from a fixed distribution nor independent and identically distributed. Instead, the distribution of the observations $\{o\}$ depends on the current policy for choosing actions and the chosen action affects the probability of the next observation as well.
Besides, although shortest paths as the optimal trajectories could be generated and made accessible under simulation platforms, it is expensive and almost impractical to ascertain in real world scenarios. Moreover, for some visual navigation tasks, the optimal trajectories are unavailable and even unattainable. For example, the optimal trajectories of VLN task are generated from human annotations. In Embodied Question Answering task, instead of the shortest paths, the optimal trajectories should help the agent to collect useful observations to answer the user-specified questions correctly. Similarly in the Embodied Amodal Recognition task, the optimal trajectories are the ones to assist the agent to recognize the occluded objects as early as possible. 
As a result, researchers formulate the VIN problem as a (Partially Observable) Markov Decision Process (MDP) and address it within the deep reinforcement learning paradigm.

In the MDP setting, the agent's visual input $o$ is defined as an observation of its hidden state. At each time step, the agent takes an action $a$ to transit from its current state to a new state which yields a new observation, and then receives rewards as feedback until it reaches a goal state $g$. The agent collects experience through its trial and error interactions with the environment and learns the optimal action policy $P(a|o;g)$ by maximizing the expected cumulative rewards. 

While the optimal trajectories as supervisions are not required, solving VIN under the MDP setting heavily relies on sample efficiency.

A straightforward way to achieve high sample efficiency is to imitate experts' policies, such as the ones generating the shortest and the annotated paths \cite{das2018embodied,das2018neural,wang2019reinforced,hong2021vln}. After pre-trained on the samples collected by following the experts, the models are further fine-tuned with reinforcement learning algorithms. The other way to improve sample efficiency is to define a denser reward function. The authors of \cite{ye2018active,ye2019gaple} defined the reward based on the size of the bounding box of the target object from the agent's detection system in order to solve their object search task. A much denser reward function was defined in \cite{wang2018look,tan2019learning,huang2019transferable,wu2019bayesian}, where they calculated the change in distance to the goal location as the immediate reward for the performed action. Such a dense reward function was also applied in \cite{das2018embodied,das2018neural,yu2019multi,li2019walking} to help the agent get close to the goal locations. Additionally, they adopted the accuracy of their question answering models as the final reward in order to perform the EQA task well. In the absence of the dense reward function, \cite{ye2021efficient} presented an efficient exploration strategy enabled by a hierarchical reinforcement learning paradigm to improve the sample efficiency.


Efforts have also been made towards capturing meaningful state representations so that their models can be learned with fewer samples. \cite{zhu2017target} adopted siamese layers to capture spatial arrangement between the agent's visual observation and the goal observation. \cite{ye2019gaple} represented the hidden state with the semantic masks and depth information estimated from the visual observation. \cite{wu2019exploring} proposed an inverse dynamic model to capture the state representation by predicting the action given two adjacent visual observations in a self-supervised manner. \cite{yang2018visual,li2019walking,du2020learning} augmented the visual observation with additional information to regularize the learning of the state representations. 
 In \cite{yang2018visual}, the authors supplemented the visual observation with the object relational graph extracted from the Visual Genome \cite{krishna2017visual} corpus to perform their object navigation task. Similar in \cite{du2020learning}, the authors instead built the object relational graph from the agent's exploration experience in the training environments.
The authors of \cite{li2019walking} incorporated the predicted next observations into the state representations. 

\subsection{The Holy Grail: Generalization}
While generalization ability is always adopted to evaluate learning-based approaches, especially the supervised learning methods, the definition of generalizing in VIN is still an open problem due to varying settings. In general, the generalization ability in the visual navigation task denotes how well a trained visual navigation model performs on an unseen environment to achieve a new homogeneous user-specified goal. When a model is trained with optimal trajectories under the seen environments in a supervised way, it is natural to evaluate its generalization ability without an extra training process, as most of the work we enumerate in Section~\ref{sec:sl} did. However, when viewing the visual navigation task as the MDP problem and solving it with deep reinforcement learning (DRL), the methods are unlikely to generalize well to unseen environments or goals, given the reinforcement learning methods are designed to tackle a fixed MDP problem on seen environments. Unseen environments and goals typically define a set of brand new MDP problems that require an additional learning process.

Several methods allowed additional training steps on unseen environments to improve their generalization ability. \cite{zhu2017target} designed scene-specific layers to enable their trained models to generalize to unseen environments with a much smaller number of extra training iterations. The authors of \cite{wortsman2019learning} adopted a meta-learning-based method to quickly adapt the learned model to the new environments with the proposed self-supervised interaction loss. A few training steps are required by \cite{savinov2018semi,gupta2017cognitive,chaplot2020object} to build the maps for the new environments before performing the VIN tasks.

More research work focused on learning generalizable representations from the agent's observation $o$ and the goal description $g$ and with which the derived action policy can generalize to new environments without extra training steps. \cite{ye2019gaple} and \cite{mousavian2019visual} showed that the semantic segmentation and the depth map can help improve the generalization ability. \cite{yang2018visual} improved the generalization ability by embedding an object relational graph learned from the Visual Genome dataset \cite{krishna2017visual} into the state representation. \cite{wu2019bayesian} and \cite{ye2021hierarchical} adopted a probabilistic graph to capture the room layout and the object layout respectively. The graph can be taken as a high-level planner integrated with reinforcement learning based locomotion policy in order to achieve better generalization performance.

Data augmentation has also been explored to improve VIN generalization ability. For instance, the Speaker-Follower models introduced in \cite{fried2018speaker} augmented data by adopting its speaker model to create synthetic instructions on sampled new routes. \cite{tan2019learning} came up with the ``environment dropout'' method to mimic unseen environments. \cite{parvaneh2020counterfactual} generated counterfactual environments to account for unseen scenarios. \cite{devo2020towards} decoupled the object localization network from the navigation network so that the domain randomization techniques can be effectively applied on the object localization network only. \cite{maksymets2021thda} augmented the set of training layouts by inserting 3D scans of household objects at arbitrary scene locations.

\section{Further Discussions}
The goal of the visual navigation task is to equip an intelligent agent with the capability of navigating towards a user-specified goal under any environments, especially under real-world environments. To this end, we discuss certain critical issues remaining unsolved in the recent work as summarized in Section~\ref{sec:summary}.

\subsection{Simulation vs Real-world}
Simulation platforms allow almost unlimited experiments without the hassles of dealing with the time-consuming mechanical work on a physical robot, which largely improves the data efficiency and facilitates the research on the VIN task. To reduce the gap between the simulation and the real-world environment, some simulation platforms, like the R2R dataset \cite{anderson2018vision}, build the virtual environments upon real images. However, simulations are still far away from the real word as a large chunk of uncertainties in the real world cannot be captured and accurately modeled, impeding the transfer of the progress achieved on simulation platforms to the real-world scenarios.

One benefit of the simulation platforms is the feasibility of generating the shortest paths as the supervised signals to train visual navigation models for certain tasks. However, such shortest paths are typically generated without taking the real-world's uncertainty into consideration. One significant factor among them is the physical robots' control errors. Thus, whether the feature representations learned from the oracle shortest paths in the simulation platforms can adapt to the real-world environments remains unclear. We argue that achieving this is critical, as generating shortest paths in real-world environments is extremely expensive, and even generating a small number of samples for model fine-tuning may not be affordable. In addition, even in the real image constructed simulated environments, the environments are static, far away from the reality where the real environments are subject to change of light, objects layout etc. Therefore, the solutions under the static simulated environments still leave much to be desired, make it challenging to explore the visual navigation task under dynamic environments.

As also pointed out by \cite{kadian2019we}, the progress on simulation platforms does not hold well in reality, since the virtual agents tend to take advantage of the simulators imperfection. In their point goal navigation experiments conducted on Habitat \cite{savva2019habitat}, they found the virtual agents are able to slide around the obstacles to reach a desirable state which would not happen in the real world. In other simulation platforms and experiments \cite{zhu2017target,ye2018active}, the agents simply stay at the current position without changing the environment when a collision happens, which is also unrealistic. To this end, many works study the simulation to real transfer in the visual navigation domain \cite{bharadhwaj2019data}, while in \cite{kadian2019we}, the authors suggest to evaluate simulators in terms of how likely the performance improvement achieved on them can hold in reality, rather than their visual or physical realism. To conclude, the gap between simulation and real-world still needs to be further studied with caution on claims being made that are validated on simulated environments only.

\subsection{Supervised vs Reinforcement Learning}
As described in Section~\ref{sec:summary}, there are two primary methods being used in VIN tasks. One is the supervised learning method of matching the predicted actions to the available ground-truth navigation trajectories and the other is the reinforcement learning method of solving the VIN tasks by maximizing the user-defined rewards. Though the efficacy of both methods has been demonstrated and they are usually used simultaneously, the applicability and the respective limitations have not been explicitly made clear.

While the supervised learning method is straightforward yet powerful, it requires a large amount of ground-truth annotations as supervised signals. For certain tasks, such as the point goal navigation, it makes sense to take the shortest paths as the ground-truth trajectories. Typically, it is easy to generate the shortest paths in terms of the agent's available actions under simulation platforms when totally ignoring the control noises. However, in real-world scenarios or simulation platforms that model uncertainties of the real world, generating the shortest paths is expensive, and thus limits the use of the supervised learning method. Moreover, the shortest paths are not always optimal or ground-truth trajectories for other tasks, such as the EQA, VLN and EAR tasks. The optimal trajectories are either not accessible or require intensive labor, making the supervised learning methods not applicable. More importantly, the supervised learning methods might be fragile because a prediction error would accumulate fast in the sequential process of the VIN tasks. On the contrary, the reinforcement learning method is always applicable without the requirement of the existence of ground-truth trajectories. However, its sample inefficiency issue makes the learning extremely difficult especially when the inputs are of high dimension.

Additionally, even in the situations where both supervised learning method and reinforcement learning method are applicable, it is not fair to directly compare the two methodologies. Firstly, the supervised learning method requires optimal trajectories under the training environments as additional information compared to the reinforcement learning method. Secondly, the goal of the supervised learning methods is to learn the feature representations with the optimal trajectories under the training environments that can generalize to the testing environments, where the training and testing environments are homogeneous. Therefore, the evaluation of the supervised learning method is typically conducted on the testing environments where no extra training process is allowed. In comparison, the reinforcement learning method assumes the optimal trajectories are unknown under the training environments, and its ultimate goal is to figure out the optimal trajectories through many trial and error interactions with these environments. As a result, the reinforcement learning method is typically evaluated in the training environments or on a testing environment that allows further training process.

In summation, the supervised learning method is more generalizable in homogeneous tasks, but its applicability is limited by the availability of the ground-truth solutions. The reinforcement learning method carries less generalization ability, but the methodology applied on one task sheds light on solving other tasks, including heterogeneous ones.

\subsection{Learning Objective and Generalization}

The objective of a VIN task is to learn an optimal action policy $P(a|o; g)$.  It is usually optimized towards $P(a|o;g;\hat{m})$ in the previous work whether supervised learning or reinforcement learning methods are used. Here, $\hat{m}$ represents the ground-truth map of the environment in which the agent is asked to achieve the goal $g$ with the observation $o$. In fact, $P(a|o;g) = \sum_{m}P(a|o;g;m)P(m|o;g)$ where $m$ denotes any one of all the possible maps. Therefore, the learning objective is correct if and only if $P(\hat{m}|o;g)=1$, which means the ground-truth map $\hat{m}$ should be deterministic given the observation $o$ and the goal description $g$. However, for most VIN tasks where the goal $g$ is far away and cannot be grounded in the observation $o$, the assumption does not hold and the learning objective would make the model memorize the environment maps using spurious features and thus lose the generalization ability on the new environments. Such the overfitting problem is alleviated in the step-by-step VLN task since the step-wise instructions $g$ are more likely to be grounded in each observation $o$ to determine the optimal action $a$. However, it is not resolved because the agent might deviate from the ground-truth path and ultimately make the instructions no longer be able to be grounded.

The aforementioned fact actually suggests several ways to improve the generalization ability of a VIN model. First, the training data can be augmented to have the same probability distribution $P(m|o;g)$ as the testing data. Existing datasets used for VIN tasks \cite{kolve2017ai2,anderson2018vision} are small and static, and the environment maps $m$ and goal locations $g$ are always deterministic given any one of the observations from the environments. It is not true in reality and may not hold in testing data. Therefore, augmenting the training data by changing the goal locations and modifying the environment maps should be helpful.

Second, since the learning objective makes more sense when there is less uncertainty about the environment map given the inputs, we can supplement additional information to reduce the uncertainty, such as the history experiences $h$ collected by the agent  \cite{chen2021history}. Specifically, instead of learning $P(a|o; g)$, we can learn $P(a|o;g;h)$ because $P(\hat{m}|o;g;h) > P(\hat{m}|o;g) $ makes the learning objective more accurate.

Last, the probability distribution $P(m|o;g)$ could be modeled explicitly to improve the VIN generalization ability. While a straightforward way is to build the environment map $m$ from the observation $o$ as the ones presented in \cite{savinov2018semi,gupta2017cognitive,chaplot2020object}, these low-level map representations are environment-dependent and thus become useless when encountering a new environment. Instead, a universal high-level environment representation like the one proposed by \cite{ye2021hierarchical} is preferable. Moreover, it is also promising to infer the probability distribution $P(m|o;g)$ from external knowledge, such as human-provided instructions and public knowledge bases.

\subsection{The Role of Knowledge}

The importance of knowledge has long been identified in many AI tasks, such as image recognition \cite{aditya2019integrating}. Intuitively, knowledge should also be of great benefit to better perform the VIN task. Even for human beings, it is much easier to navigate in a structured indoor environments than a contextless maze, indicating that the high navigation performance is usually achieved by reasoning from observations rather than merely memorizing the environments. To be specific, the indoor environments typically have distinct structures, such as functional areas and object layouts in houses. With the knowledge of such structure, the agent is expected to explore the environment more efficiently by avoiding getting trapped at irrelevant locations. For example, with the commonsense knowledge that a sofa is typically found in the living room, the agent shouldn't spend much time in the kitchen to find a sofa. Moreover, such knowledge is likely to still hold in previously unseen environments, making it possible to achieving better generalization ability.


A growing number of works start to exploit knowledge to help perform the visual navigation task, typically the room and/or object navigation tasks where the semantic labels of the rooms and/or objects are specified as the navigation goals (Section.~\ref{sec:summary_rep}). 
In \cite{yang2018visual}, the authors extracted objects relations from the Visual Genome \cite{krishna2017visual} corpus and incorporate this prior into their models through Graph Convolutional Networks. 
 The authors of \cite{wu2019bayesian} came up with the Bayesian Relational Memory (BRM) architecture to capture the room layouts of the training environments from the agent's own experience for room navigation. 
More generally, \cite{ye2021hierarchical} captured underlying relations among all goals in the goal space to guide the goal-driven visual navigation tasks.
Unarguably, these approaches are still at its preliminary stages comparing with how human beings perform the VIN task, indicating a fruitful direction to be further explored.

\section{Summary and Future Work}
In this paper, we discussed the recent advances in learning-based visual indoor navigation. We first summarized all relevant tasks into three categories in terms of the descriptions of the user-specified goals, i.e. label indicated goals, image indicated goals, and language indicated goals.
We further described the two primary methods that are applied under certain conditions. One is the supervised learning method aimed to learn generalizable feature representations given the optimal trajectories are available or easy to obtain. The other is the reinforcement learning  method of taking the visual navigation task as a MDP problem. Since the reinforcement learning method is not as suited for generalization as the supervised learning method is, we also introduced existing studies on improving its generalization ability for the VIN task. 

To help further performance improvements in VIN , we pointed out several issues with the current field which are worth future studies:
1) the legality of studying the VIN task on the simulation platforms; 2) the fairness in comparing the supervised learning based methods and the RL based ones; 3) the potential for improving the generalization ability of VIN models; 4) the avenue of integrating knowledge to guide the learning process tackling VIN.

\newpage
{
\bibliographystyle{named}
\bibliography{ijcai22}
}

\end{document}